\documentclass[10pt,twocolumn,letterpaper]{article}

\usepackage{iccv}              
\usepackage{algorithm}
\usepackage{algpseudocode}
\usepackage{graphicx, multirow, booktabs}
\usepackage{upgreek}
\usepackage{lineno}  

\definecolor{iccvblue}{rgb}{0.21,0.49,0.74}
\usepackage[pagebackref,breaklinks,colorlinks,allcolors=iccvblue]{hyperref}

\def\paperID{*****} 
\def\confName{ICCV}
\def\confYear{2025}

\title{Harnessing SPAD Noise: \\Test-Time Gated Attention for Indoor Depth Estimation}

\author{Yiquan Gao\\
Heriot-Watt University\\
}

\begin{document}
\maketitle
\begin{abstract}
	Monocular depth estimation (MDE) has emerged as a cost-effective alternative to hardware-dependent 3D sensing modalities, yet predicting precise metric depth remains a persistent challenge due to intrinsic scale ambiguity, edge bleeding, and limited zero-shot generalization in unseen environments. Recent multi-modal approaches incorporate 1D/3D Single-Photon Avalanche Diode (SPAD) data to resolve these ambiguities, but they mainly treat SPAD noise as a harmful artifact required for denoising or leave the noise ignored. In this paper, we present an essentially different insight: native SPAD noise is an informative prior that implicitly encodes deterministic environmental characteristics. Driven by this paradigm shift, we introduce a novel, lightweight Test-Time Gated Attention framework by Harnessing SPAD Noise, therefore dubbed HSN. Tailored for indoor scenarios, HSN leverages the synergistic combination of 1D SPAD histograms and Gate Adapters (GAs). Through three complementary optimization phases, it generates channel-wise gated attention vectors to dynamically adjust encoder feature maps at inference time, effectively resolving ambiguous scale and boundary challenges without the heavy computational overhead or dense 3D modalities that other TTA methods suffer from. Extensive experiments on challenging real-world indoor benchmarks show that HSN achieves state-of-the-art metric depth accuracy in in-domain settings, while exhibiting even greater potential under zero-shot generalization to unfamiliar environments. Code will be made publicly available.
\end{abstract}
    
\section{Introduction}
\label{sec:intro}
Although hardware-based 3D sensing modalities like LiDAR, stereo vision, time-of-flight, and structured light are widely used, monocular depth estimation (MDE) has emerged as a compelling solution requiring only a single camera. By eliminating expensive and bulky hardware dependencies, MDE provides a versatile alternative for diverse fields, including indoor navigation~\cite{zheng2024monocular}, robotics~\cite{dong2022towards}, augmented reality~\cite{ganj2024mobile}, and AI-generated content (AIGC)~\cite{lyu2026depth}. While MDE excels at capturing relative depth cues, predicting precise metric depth (i.e., absolute distances) remains a persistent bottleneck. This limitation primarily stems from the scale ambiguity intrinsic to monocular vision, alongside severe edge bleeding at boundaries and structural artifacts within object surfaces. Furthermore, these challenges become seriously exacerbated during zero-shot generalization in unseen environments.

To tackle these issues, recent works have incorporated various modalities, such as 3D SPAD volumes~\cite{lindell2018single,sun2020spadnet}, 3D point clouds~\cite{marsal2025simple}, IMU~\cite{yang2025zero}, and 1D SPAD histograms~\cite{nishimura2020disambiguating,lee2023monocular}, focusing on either multimodal feature fusion during full training or rescaling at test-time adaptation. Compared to other modalities, 1D SPAD histograms are acquired via a single-pixel time-resolved detector with a low-power pulsed light source, which renders them the most cost-effective yet highly noisy data form, offering the least amount of information. Confronted with SPAD noise, existing 3D SPAD depth networks~\cite{lindell2018single,sun2020spadnet,peng2020photon} predominantly focus on denoising to eliminate these fluctuations as destructive artifacts. Meanwhile, current 1D SPAD methods either integrate the histograms into deep multimodal feature fusion during full training~\cite{lee2023monocular}, or restrict their utility to test-time histogram matching~\cite{nishimura2020disambiguating}. As a result, they completely bypass raw noise processing, leaving the statistical properties of the 1D SPAD noise unexploited.

Departing from all these existing tracks, we present a fundamentally different perspective: this native noise is not merely a pure corruption, but rather an informative prior that implicitly encodes deterministic environmental characteristics. To realize this, our work introduces a novel Test-Time Gated Attention framework named HSN that explicitly harnesses the SPAD noise prior inherently embedded within 1D SPAD histograms. Specifically, this framework aims at refining depth perception for indoor scenarios during test time, since 1D SPAD signals suffer from severe ambient light interference outdoors but maintain a relatively stable noise structure indoors. In the indoor setting, HSN uses the noise as a guidance signal to yield a channel-wise gated attention vector that dynamically calibrates the corresponding encoder maps, thereby effectively addressing the ambiguous scale and boundary challenges in monocular metric depth estimation. By building upon the efficient combination of 1D SPAD and MLPs, HSN achieves a highly lightweight TTA solution that thrives under extremely sparse data regimes, which avoids heavy TTA designs such as the diffusion overhead of MfH~\cite{zhao2024metric} and the reliance on dense 3D point clouds in DR-LiDAR~\cite{marsal2025simple}.

Our main contributions can be summarized as follows:
\begin{itemize}
	\item \textbf{A Novel Noise-Driven Paradigm:} We challenge the convention of viewing SPAD noise solely as a destructive artifact destined for denoising. Instead, we introduce a paradigm shift by repurposing the 1D SPAD noise signal as an informative prior that correlates with environmental characteristics to enable geometric and scale calibration for indoor monocular metric depth estimation.
	\item \textbf{Lightweight Test-Time Adaptation Framework:} We propose an efficient framework named Test-Time Gated Attention by Harnessing SPAD Noise (HSN), along with the design of three complementary optimizations: statistical distribution alignment, physically-constrained losses, and a stability audit mechanism. By exploiting the synergy between 1D SPAD histograms and Gate Adapters, HSN yields channel-wise gated attention vectors to adaptively calibrate encoder feature maps at inference time, eliminating the heavy computational overhead and dense 3D modalities common in competitive TTA methods.
	\item \textbf{Superior In-Domain and Zero-Shot Generalization:} As a generic plug-and-play TTA framework for encoder--decoder MDE architectures, HSN is validated via extensive experiments on challenging indoor benchmarks involving real-world data. While achieving state-of-the-art metric depth accuracy for the in-domain setting, it exhibits even more profound potential in zero-shot generalization across unseen scenes. 
	\end{itemize}

\section{Related Work}
\label{sec:relatedwork}

\noindent\textbf{Monocular Metric Depth Estimation.} 
Conventional absolute depth perception methods heavily rely on stereo vision or expensive active sensors like LiDAR to resolve real-world dimensions. To circumvent these hardware constraints, recent deep neural networks formulate Monocular Metric Depth Estimation (MMDE) as a direct mapping from single RGB images to physical distances, typically by fitting unknown camera intrinsics onto canonical camera spaces. Despite notable progress on specific datasets, existing MMDE networks, including regression models~\cite{eigen2014depth,fu2018deep}, discrete bin classification models~\cite{bhat2022localbins,bhat2021adabins} and depth foundation models~\cite{bhat2023zoedepth,yang2024depth,yang2024depthv2}, still struggle with scale ambiguity~\cite{hartley2003multiple,eigen2014depth} when transferring global scene scales to unseen, zero-shot scenarios with varying camera intrinsics and environmental factors. Multimodal feature fusion~\cite{sun2020spadnet,lee2023monocular,zhang2022can,zhu2024scaledepth}, large-scale data mixture~\cite{ranftl2020towards,yin2023metric3d}, and camera-depth disentanglement~\cite{piccinelli2024unidepth,piccinelli2025unidepthv2} enable training models across multiple modalities or domains simultaneously, which significantly extends the generalization capabilities for zero-shot MMDE. Nevertheless, when encountering exceptionally challenging open-world domains, these models commonly require targeted finetuning to resolve severe out-of-distribution gaps due to their lack of test-time scale alignment. However, downstream finetuning is both computationally prohibitive and highly inflexible for practical deployment. Therefore, our HSN provides a general plug-and-play TTA framework for various categories of MMDE networks, contingent upon their utilization of an encoder-decoder architecture.

\noindent\textbf{Test-Time Adaptation for Depth Perception.} 
Deploying a pre-trained MMDE model into unobserved domains often induces massive performance degradation due to distribution shifts in sensor intrinsics, illumination conditions, and environmental variations. To alleviate this without re-triggering expensive offline finetuning, Test-Time Adaptation (TTA) focuses on dynamically adjusting model weights or recalibrating output distributions during the inference stage. Although TTA for zero-shot MMDE remains heavily under-explored, the few existing approaches mainly leverage cross-modal visual clues or domain-specific priors to bridge domain gaps. For instance, MfH~\cite{zhao2024metric} adopts a 'generate-and-estimate' paradigm via diffusion-based human inpainting, while SurgDepth~\cite{wang2026surgdepth} adapts pre-trained decoders exclusively for medical endoscopy using surgical self-supervised signals. However, these methods either introduce a heavy generative loop or require repetitive online optimization, incurring severe computational bottlenecks during TTA. Another line of work, such as DR-LiDAR~\cite{marsal2025simple}, utilizes hardware-dependent dense 3D points from sensors to rescale an affine-invariant disparity map, yet its requirement for explicit 3D coordinates leads to substantial computational and financial overhead, restricting its deployment on low-cost commodity robots. In contrast, we present a cost-effective TTA design by leveraging extremely sparse 1D SPAD inputs coupled with a compact MLP architecture, ensuring a versatile solution with highly competitive performance for robust metric depth adaptation in zero-shot MMDE.

\noindent\textbf{Exploiting Noise in Learning.} 
Noise can manifest in various forms and is generally regarded as a harmful artifact that needs to be removed in numerous applications, such as license plate detection~\cite{gao2018implementation}, EEG analysis~\cite{lai2018artifacts}, point cloud denoising~\cite{zhou2022point}, and time series forecasting ~\cite{zhang2026fusad}. This denoising practice dominates the Single-Photon Avalanche Diode (SPAD) field, since raw SPAD acquisitions frequently suffer from unavoidable environmental interference (e.g., clutter from complex background objects and unpredictable ambient light especially outdoors) and intrinsic sensor disturbances like dark count rates (DCR). Consequently, SPAD-based MDE methods~\cite{lindell2018single,sun2020spadnet,peng2020photon} prioritize raw 3D dense volume denoising as an indispensable prerequisite. However, existing studies regarding the noise handling of the 1D sparse SPAD representation remain absent~\cite{nishimura2020disambiguating,lee2023monocular}, leaving its significantly higher noise completely untouched.

Contemporary works like BadCLIP~\cite{bai2024badclip} and BAPLE~\cite{hanif2024baple} emerge to show how learnable noise can backdoor VLMs, TNT~\cite{imam2025noise} shifts this utility toward TTA for zero-shot image classification, proving that such noise-based optimization within input space can serve as an efficient tool to handle unpredictable distribution shifts. Inspired by this philosophy, we present a distinct insight against conventional SPAD processing: rather than a destructive artifact to be filtered out, native SPAD noise constitutes an informative prior that implicitly encodes deterministic environmental characteristics. To this end, we propose the Test-Time Gated
Attention framework by harnessing SPAD noise to address zero-shot MMDE, marking the first attempt to exploit noise learning for test-time depth adaptation. 

\section{HSN: Test-Time Gated Attention}
\label{sec:Methodology}

\begin{figure*}[t]
	\centering
	\includegraphics[width=\linewidth]{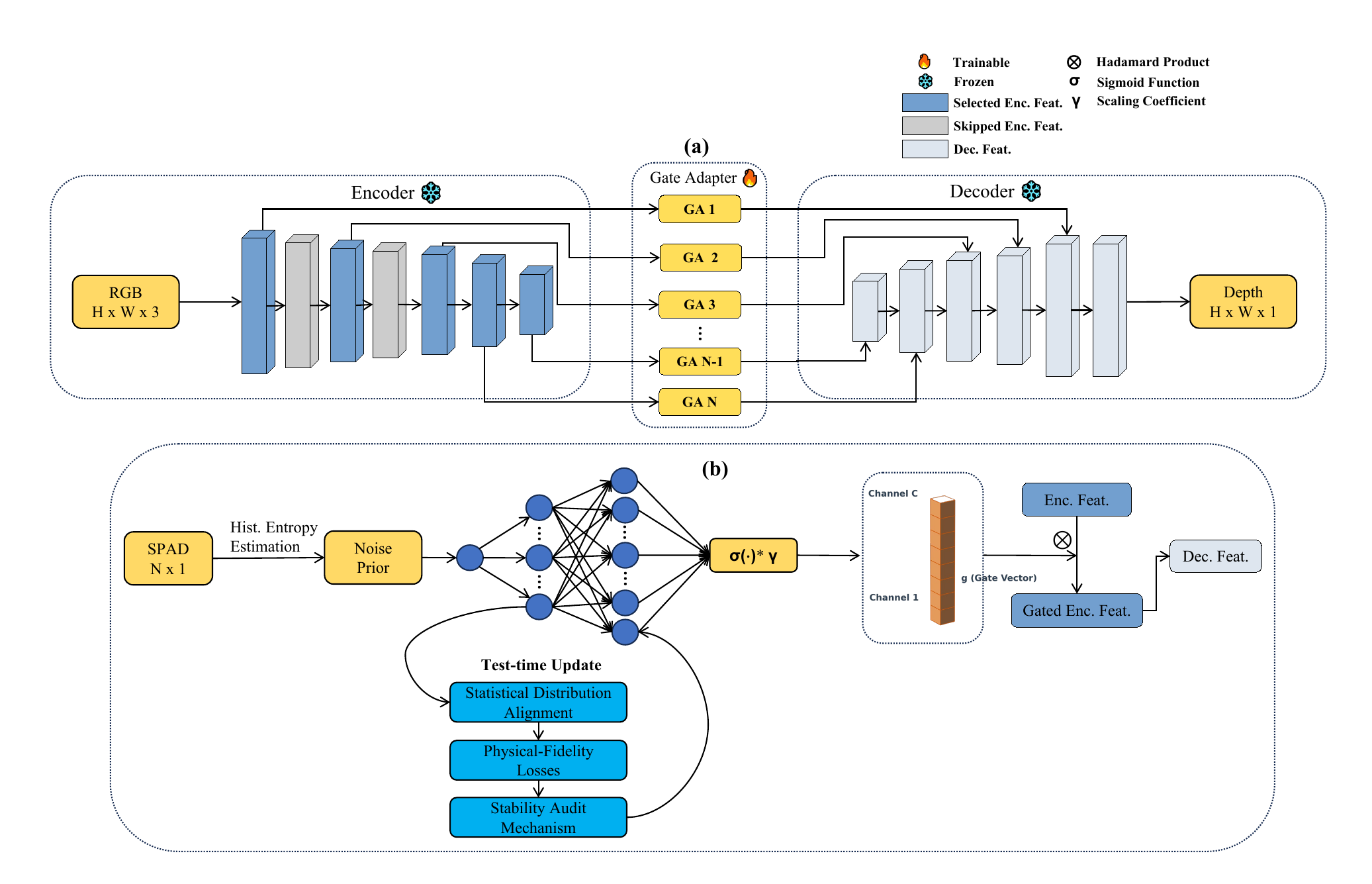}
	\caption{\textbf{Overview of the proposed Test-Time Gated Attention framework (HSN).} (a) illustrates how Gate Adapter (GA) modules are embedded between the selected encoder map and its corresponding decoder map, where the targeted encoder map is routed to the subsequent decoding layer. Conversely, encoder maps not routed to decoding layers bypass GA modulation entirely. (b) details the architectural design of the GA module and the TTA workflow. During inference, the pre-trained encoder-decoder network $\theta$ remains frozen to preserve its inherent structural priors, restricting optimization strictly to the GA parameters $\phi$. These modules leverage the SPAD noise signal to dynamically generate a channel-wise gated attention vector $g$, thereby calibrating the selected encoder maps prior to decoding.}
	\label{fig:hsn_flowchart}
\end{figure*}

 We propose a Test-Time Gated Attention framework dubbed \textbf{HSN} to refine monocular depth predictions by \textbf{H}arnessing \textbf{S}PAD \textbf{N}oise, as illustrated in Fig.~\ref{fig:hsn_flowchart} and detailed in Alg.~\ref{alg:tta_process}. During inference, the pre-trained encoder-decoder network $\theta$ remains frozen to preserve its general structural priors. HSN only optimizes a specific subset of parameters within the Gate Adapter (GA) modules $\phi$ at test time. This test-time adaptation (TTA) process comprises three distinct phases: statistical distribution alignment, physically-constrained optimization, and a robustness audit to ensure reliability. Before dissecting these three phases, we first elaborate on the structural design of the GA modules and the overall  workflow that governs their online adaptation.

\subsection{Gate Adapter and TTA Workflow}
\label{subsec:Gate Adapter and TTA Workflow}
As shown in Fig.~\ref{fig:hsn_flowchart}(a), the depth estimation network $\mathcal{F}_{\theta, \phi}$ maps an input RGB $I \in \mathbb{R}^{H \times W \times 3}$ to the corresponding depth $D^* \in \mathbb{R}^{H \times W \times 1}$. This mapping is refined by embedding test-time updated GA modules, which adjust the selected encoder maps via a gated attention mechanism before the decoder layers. $\mathcal{F}_{\theta, \phi}$ is parameterized by two distinct weight sets: a frozen set $\theta$ retaining the pre-trained geometric representation of the encoder-decoder backbone, and a trainable set $\phi$ within the GA modules optimized by the HSN framework during inference. Notably, the GA modules only process the selected encoder maps that are routed directly to their corresponding decoder layers. Therefore, the parameter set $\phi$ is formalized as the collection of weights across these specific routing paths:
\begin{equation}
	\phi = \{ \phi_l \}_{l \in \mathcal{L}},
	\label{eq:phi_set}
\end{equation}
where $\phi_l$ represents the internal trainable weights of the GA module embedded at the $l$-th encoder-decoder routing pair, and $\mathcal{L}$ denotes the set of indices corresponding to the selected routing connections.

Furthermore, Fig.~\ref{fig:hsn_flowchart}(b) depicts the architecture and test-time loop for each module $\phi_l \in \phi$. Firstly, the SPAD temporal histogram $S \in \mathbb{R}^{N \times 1}$ corresponding to $I$ is converted into a noise level indicator $\omega \in \mathbb{R}$ as follows:
\begin{equation}
	\label{eq:omgega}
	\omega = \frac{-1}{\ln N + \epsilon} \sum_{i=1}^{N} p_i \ln (p_i + \epsilon)
\end{equation}
where $p_i = S_{i} / (\sum_{k=1}^{N} S_{k} + \epsilon)$ represents the normalized temporal probability of the $i$-th bin, $N$ denotes the total number of temporal bins, and $\epsilon = 10^{-9}$ ensures numerical stability. Under this formulation, the noise scalar is strictly bounded within the interval $\omega \in [0, 1]$. Intuitively, a value close to 1 indicates a noise-dominant histogram with scattered background photons, whereas a value approaching 0 represents a sharp, clean signal peak.

Subsequently, the noise scalar $\omega$ serves as an auxiliary input to each GA module $\phi_l$:
\begin{equation}
	F_{out} = F_{in} \otimes \left( \upgamma \cdot \sigma\left( \text{MLP}(\omega) \right) \right)
\end{equation}
where $F_{in}, F_{out} \in \mathbb{R}^{C_l \times H_l \times W_l}$ denote the input and output spatial features of the $l$-th module respectively. Channel-wise gated attention vector $g$ is the output of $\upgamma \cdot \sigma\left( \text{MLP}(\omega) \right)$, where $\sigma(\cdot)$ represents the Sigmoid function, $\upgamma$ is a scaling coefficient, and $\otimes$ denotes the Hadamard product with broadcasting. For notation simplicity, we omit the index $l$ for the internal parameters of $\text{MLP}(\cdot)$. To ensure test-time efficiency, it adopts a lightweight bottleneck design ($1 \to \frac{C_l}{4} \to C_l$):
\begin{equation}
	\text{MLP}(\omega) = W_2 \cdot \text{ReLU}\left(W_1 \cdot \omega + b_1\right) + b_2
\end{equation}
where $W_1 \in \mathbb{R}^{\frac{C_l}{4} \times 1}$, $W_2 \in \mathbb{R}^{C_l \times \frac{C_l}{4}}$, $b_1 \in \mathbb{R}^{\frac{C_l}{4}}$, and $b_2 \in \mathbb{R}^{C_l}$.

Crucially, the final parameters $\{W_2, b_2\}$ are zero-initialized, ensuring $\text{MLP}(\omega) = 0$ regardless of the input $\omega$ at the inception of TTA. By setting $\upgamma = 2$, the initial modulation factor becomes strictly $2 \cdot \sigma(0) = 1$, guaranteeing that the GA module fully preserves the pre-trained baseline features at the start of inference. As TTA progresses, this mechanism adaptively scales the feature maps within a flexible range of $(0, \upgamma)$, which dynamically modulates the spatial feature representations conditioned on the temporal noise statistics derived from the SPAD histogram, thereby effectively calibrating the model's sensitivity and robustness to geometric variations while mitigating depth ambiguity. During adaptation, only the parameters of $\phi_l$, specifically $\{W_1, W_2\}$ and $\{b_1, b_2\}$, are optimized across three sequential phases: statistical distribution alignment, physically-constrained optimization, and stability auditing, which are described below.

\subsection{Statistical Distribution Alignment}
\label{subsec:Statistical Distribution Alignment}
Given an input image $I$ and its corresponding SPAD temporal histogram $S$, we first perform an initial forward pass to obtain $D_{init} = \mathcal{F}_{\theta, \phi}(I, S)$. To address the systematic domain gaps often encountered in unseen data, we introduce a Zero-order Distribution Alignment (ZoDA) scheme. 

We first derive the physical depth expectation $\mu_s$ from the SPAD histogram $S$:
\begin{equation}
	\label{eq:mu_s}
	\mu_s = \frac{\sum_{i=1}^{B} S_i \cdot z_i}{\sum_{i=1}^{B} S_i + \epsilon_s}
\end{equation}
where $S_i$ denotes the photon count of the $i$-th bin, and $z_i$ represents the corresponding depth value uniformly sampled from the depth interval $[d_{min}, d_{max}]$. $\epsilon_s$ is a small constant for numerical stability. The initial prediction is then refined via a damped global offset to align its distribution center with the physical prior $\mu_s$:
\begin{equation}
	\label{eq:distribution_align}
	D_{base} = \text{clamp}(D_{init} + \kappa \cdot (\mu_s - \bar{D}_{init}), d_{min}, d_{max})
\end{equation}
where $\bar{D}_{init}$ is the spatial mean of $D_{init}$, and $\kappa = 0.65$ is an empirical attenuation factor for unseen data. For in-domain data where the initial geometry is reliable, we set $\kappa = 0$ to preserve the original scale. 

An adaptive constraint modulation strategy is applied to the hyperparameter set $\mathcal{P} = \{\lambda_f, \gamma, \epsilon\}$, dynamically updating the parameters across both unseen and in-domain scenarios to maintain an optimal equilibrium between aggressive correction and structural preservation.
For unseen data, we enforce a conservative adaptation trajectory by assigning a fixed, high-intensity fidelity weight $\lambda_f = 1.5 \times 10^6$, a stringent reliability threshold $\gamma = 0.995$, and a restrictive gradient clipping threshold $\epsilon = 0.02$. This configuration ensures that the optimization remains heavily anchored to the aligned baseline, preventing structural distortions in unfamiliar domains.

Conversely, for in-domain data, we introduce a noise-aware scaling to facilitate precision refinement, which is inspired by the data uncertainty concept in illumination enhancement \cite{gao2026data}. Specifically, the fidelity weight $\lambda_f$ is adaptively scaled according to the noise level $\omega$ of the SPAD histogram in Eq.~\ref{eq:omgega} using an exponential kernel:
\begin{equation}
	\label{eq:adaptive_scale}
	\lambda_f = \text{clamp}(C_1 \cdot \exp(\alpha \cdot \omega), \lambda_{min}, \lambda_{max})
\end{equation}
where $C_1 = 4 \times 10^5$ and $\alpha = 2.5$ are empirically set to modulate the exponential sensitivity to $\omega$ while balancing numerical stability. This formulation allows the optimizer to perform surgical residual corrections in high-SNR (Signal-to-Noise Ratio) scenarios while remaining conservative in the presence of heavy ambient noise. To accommodate these subtle refinements, the reliability threshold is relaxed to $\gamma = 0.998$ and the gradient clipping threshold is elevated to $\epsilon = 0.05$, granting the optimizer sufficient flexibility to polish the final depth map. 

Furthermore, we perform a pre-optimization evaluation by computing the initial loss $\mathcal{L}_{init}$ using the physical loss function in Sec.~\ref{subsec:Physically-Constrained Optimization}. If $\mathcal{L}_{init} < \tau_{exit}$, indicating the initial state is already physically consistent, an early exit is triggered to directly yield $D^*$ via $\Psi(\cdot)$ in Eq.~\ref{eq:post-processing}, thereby bypassing redundant iterations.

\subsection{Physically-Constrained Optimization}
\label{subsec:Physically-Constrained Optimization}
To adapt the model $\mathcal{F}_{\theta, \phi}$ to the SPAD signal characteristics and the spatial geometry between the input RGB $I$ and output depth $D_{base}$, we devise a joint physical-fidelity objective $\mathcal{L}_{TTA}$ and minimize it over $T$ iterations to fine-tune the GA parameters $\phi$:
\begin{equation}
	\label{eq:loss_tta}
	\mathcal{L}_{TTA} = \omega \cdot (\lambda_{p} \mathcal{L}_{phy} + \lambda_{f} \mathcal{L}_{fid})
\end{equation}
where $\mathcal{L}_{phy}$ enforces consistency with the physical SPAD measurements and $\mathcal{L}_{fid}$ maintains structural fidelity relative to $I$ and $D_{base}$. $\lambda_p = 0.05$ and $\lambda_f \in \mathcal{P}$ are loss adjusting coefficients, with $\omega$ acting as the noise-aware control.

The physical consistency loss $\mathcal{L}_{phy}$ is defined as the Poisson negative log-likelihood:
\begin{equation}\label{eq:poisson_loss}
	\mathcal{L}_{phy} = \sum_{i=1}^{N} \left( \hat{\lambda}(i) - S(i) \log \hat{\lambda}(i) \right),
\end{equation}
where $S$ is the observed SPAD histogram. The estimated photon count $\hat{\lambda}(i)$ at time bin $i$ is calculated by:
\begin{equation}\label{eq:lambda_est}
	\small 
	\hat{\lambda}(i) = s_{phy} \cdot \left[ \frac{1}{H_s W_s} \sum_{j=1}^{H_s W_s} \exp \left( -\frac{(z_i - d_j)^2}{2\beta^2} \right) \right] + \epsilon,
\end{equation}
where $z_i$ is the depth value of bin $i$, $\beta = 0.03$, $\epsilon = 10^{-6}$, and $d_j$ is the $j$-th pixel value of the downsampled evolving prediction $\mathbf{d} = \downarrow(D_{tta}) \in \mathbb{R}^{H_s W_s}$. To resolve amplitude ambiguity, the stop-gradient scale factor is defined as $s_{phy} = \sum_{i=1}^N S(i) / \left(\sum_{i=1}^N \hat{\lambda}_{raw}(i) + 10^{-9}\right)$, where $\hat{\lambda}_{raw}(i)$ denotes the unscaled spatial-averaged intensity value at bin $i$ from Eq.~\ref{eq:lambda_est}.

In practice, Eq.~\ref{eq:lambda_est} acts as a virtual SPAD simulator. The Gaussian kernel mimics the temporal photon scattering caused by real-world laser pulse widths, while the spatial average simulates the low-resolution hardware characteristics. To eliminate amplitude ambiguity, $s_{phy}$ scales the unscaled probability curve to approximate the actual photon count. Guided by the Poisson distribution, minimizing $\mathcal{L}_{phy}$ in Eq.~\ref{eq:poisson_loss} enables cross-modal signal alignment, forcing the simulated and observed histograms to closely overlap.

The geometric fidelity loss $\mathcal{L}_{fid}$ is formulated to balance edge-guided spatial smoothness with global structural alignment:
\begin{equation}
	\mathcal{L}_{fid} = \lambda_{1} \mathcal{L}_{smooth} + \lambda_{2} \mathcal{L}_{huber}
\end{equation}
Here, $\mathcal{L}_{smooth}$ penalizes the absolute spatial differences of the bilinearly upsampled depth map $\hat{D}_{tta}$, where the penalty is exponentially attenuated by the absolute spatial differences of the image luminance:
\begin{equation}
	\resizebox{0.9\linewidth}{!}{%
		$\mathcal{L}_{smooth} = \operatorname{mean} \left( \Delta_x \hat{D}_{tta} \cdot e^{-\alpha \Delta_x \operatorname{gr}(I)} + \Delta_y \hat{D}_{tta} \cdot e^{-\alpha \Delta_y \operatorname{gr}(I)} \right)$%
	}
\end{equation}
Concurrently, $\mathcal{L}_{huber}$ leverages a robust Huber estimator $\text{Huber}_{\delta}(\cdot)$~\cite{loh2025theoretical} to maintain consistency between the evolving prediction $D_{tta}$ and the aligned prediction $D_{base}$ at their original resolution:
\begin{equation}
	\mathcal{L}_{huber} = \operatorname{mean} \left( \text{Huber}_{\delta} (D_{tta} - D_{base}) \right)
\end{equation}
where $\Delta_x$ and $\Delta_y$ denote the absolute horizontal and vertical first-order spatial difference operators, $\operatorname{gr}(\cdot)$ denotes the RGB-to-GRAY conversion operator, $\lambda_1 = 10^3$ and $\lambda_2 = 10^2$ are loss balancing weights, $\alpha = 20.0$ is the sharpness factor, and $\delta = 0.1$ is the Huber threshold. Functionally, $\mathcal{L}_{\text{smooth}}$ clears up depth artifacts within homogeneous object regions without blurring the sharp depth edges at geometric boundaries. Meanwhile, $\mathcal{L}_{huber}$ enforces a strict quadratic penalty on small depth variations to stabilize the global structure, but uses a robust linear penalty for large depth corrections to avoid gradient explosions during TTA. This joint objective effectively preserves sharp geometric boundaries while reinforcing global robustness against structural outliers.

During each iteration $t$, the GA parameters $\phi$ are updated using the Adam optimizer: $\phi_{t+1} = \phi_t - \eta \cdot \text{Adam}(\nabla_{\phi} \mathcal{L}_{TTA})$, where $\eta = 0.002$ is the learning rate. To ensure training stability and prevent divergence in unfamiliar domains, we apply gradient norm clipping with the threshold $\epsilon\in \mathcal{P}$ (in Sec.~\ref{subsec:Statistical Distribution Alignment}).

\subsection{Stability Audit Mechanism}
\label{subsec:Stability Audit and Post-processing}
To prevent negative adaptation, a phenomenon where test-time optimization overfits to ambient noise in unseen domains, we implement a Stability Audit (SA) mechanism. Upon completing $T$ iterations, the final physical loss $\mathcal{L}_{final} = \mathcal{L}_{phy}(D_{tta}, S)$ is verified against the initial state $\mathcal{L}_{init} = \mathcal{L}_{phy}(D_{base}, S)$ via a relative improvement criterion:
\begin{equation}
	\label{eq:relative_improvement_criterion}
	\mathcal{L}_{final} < \gamma \cdot \mathcal{L}_{init}
\end{equation}
where $\gamma$ is the scenario-dependent reliability threshold defined in Sec.~\ref{subsec:Statistical Distribution Alignment}. As specified in our modulation strategy, we apply the stringent $\gamma = 0.995$ for unseen data to mandate an assured error reduction, while the relaxed $\gamma = 0.998$ is utilized for in-domain data to permit subtle geometric refinements.

If the criterion is met, the optimized map is accepted ($D_{sel} = D_{tta}$); otherwise, the framework reverts to the aligned baseline ($D_{sel} = D_{base}$) to guarantee robustness against divergence. Finally, the selected depth map is recovered via a post-processing operator $\Psi(\cdot)$ to produce the final output:
\begin{equation}
	\label{eq:post-processing}
	D^* = \Psi(D_{sel}, I)
\end{equation}
where $\Psi$ abstracts the model-specific refinement protocols of the base model (e.g., range-clamping or spatial rescaling) to align the output prediction with the target image resolution.

\begin{algorithm}
	\caption{Test-Time Gated Attention framework (HSN)}
	\label{alg:tta_process}
	\begin{algorithmic}
		\Require Image $I$, SPAD $S$, Model $\mathcal{F}_{\theta, \phi}$, Iterations $T$
		\Ensure Optimized Depth Map $D^*$
		
		\vspace{\baselineskip}
		\State \Comment{\textbf{Sec.~\ref{subsec:Statistical Distribution Alignment}: Statistical Distribution Alignment}}
		\State $D_{init} \gets \mathcal{F}_{\theta, \phi}(I, S)$ \Comment{Initial baseline inference}
		
		\If{Data is \textbf{Unseen}} \Comment{Unseen data mode}
		\State $D_{base} \gets \text{ZoDA}(D_{init}, S)$ \Comment{Eq.~\ref{eq:mu_s}--\ref{eq:distribution_align}: Zero-order distribution alignment}
		\State Configure $\mathcal{P}_{un} = \{\lambda_{f}, \gamma, \epsilon\}$ \Comment{Set conservative modulation}
		\Else \Comment{In-domain data mode}
		\State $D_{base} \gets D_{init}$ \Comment{Maintain native distribution}
		\State Compute $\lambda_{f}$ via noise scalar $\omega$ \Comment{Eq.~\ref{eq:adaptive_scale}: Noise-aware scaling}
		\State Configure $\mathcal{P}_{in} = \{\lambda_{f}, \gamma, \epsilon\}$ \Comment{Set refinement modulation}
		\EndIf
		
		\State $\mathcal{L}_{init} \gets \mathcal{L}_{phy}(D_{base}, S)$ \Comment{Pre-optimization evaluation}
		\If{$\mathcal{L}_{init} < \tau_{exit}$} $D^* \gets \Psi(D_{base}, I)$; \Return $D^*$ \EndIf \Comment{Early exit if converged}
		
		\vspace{\baselineskip}
		\State \Comment{\textbf{Sec.~\ref{subsec:Physically-Constrained Optimization}: Physically-Constrained Optimization}}
		\For{$t = 1$ \textbf{to} $T$} \Comment{Optimization phase}
		\State Compute $\mathcal{L}_{TTA}$ via Eq.~\ref{eq:loss_tta} \Comment{Joint physical-fidelity objective}
		 \State $\phi \gets \text{Adam}(\nabla_{\phi} \mathcal{L}_{TTA}, \epsilon)$ \Comment{Update $\phi$ with gradient clipping $\epsilon$}
		\EndFor
		
		\State $D_{tta} \gets \mathcal{F}_{\theta, \phi_{new}}(I, S)$ \Comment{Adapted inference}
		\State $\mathcal{L}_{final} \gets \mathcal{L}_{phy}(D_{tta}, S)$ \Comment{Post-optimization evaluation}
		
		\vspace{\baselineskip}
		\State \Comment{\textbf{Sec.~\ref{subsec:Stability Audit and Post-processing}: Stability Audit Mechanism}}
		\If{$\mathcal{L}_{final} < \gamma \cdot \mathcal{L}_{init}$} \Comment{Reliability check}
		\State $D_{sel} \gets D_{tta}$ \Comment{Verify Eq.~\ref{eq:relative_improvement_criterion}}
		\Else \Comment{Rollback mechanism}
		\State $D_{sel} \gets D_{base}$ \Comment{Revert to aligned baseline}
		\EndIf
		
		\State $D^* \gets \Psi(D_{sel}, I)$; \Return $D^*$ \Comment{Eq.~\ref{eq:post-processing}: Post-processing}
	\end{algorithmic}
\end{algorithm}


\section{Experiments}
\label{sec:Experiments}
\subsection{Datasets}
\noindent \textbf{NYU Depth V2 (with SPAD Simulation):} This indoor benchmark is evaluated using the standard Eigen partition~\cite{eigen2014depth}, which comprises 120K training images and 654 test images of $640 \times 480$ pixels. Following Nishimura \etal~\cite{nishimura2020disambiguating}, single-pixel SPAD histograms are synthesized from dense RGB-D data by modeling pixel-wise albedo and inverse-square depth decay. The resulting macro-scale signal is then modulated by a stochastic Poisson process alongside simulated hardware jitter and background dark counts, where the signal-to-background ratio (SBR) is set to 100 and the histogram dimension $N$ is set to 1024. During depth inference, the canonical Eigen crop is applied to filter out unaligned boundary artifacts, adhering to conventional evaluation protocols. This dataset serves as a comprehensive benchmark for both in-domain and zero-shot evaluations.

\noindent \textbf{RGB-SPAD Dataset:} This real-world indoor benchmark, released by Nishimura \etal~\cite{nishimura2020disambiguating}, is employed to rigorously evaluate the practical generalization capability of different depth networks. Captured at a resolution of $640 \times 480$ pixels, it consists of aligned RGB images and authentic, hardware-captured single-pixel SPAD transient histograms from various indoor environments and target objects, paired with corresponding ground-truth depth maps. The histogram dimension $N$ is kept to 1024. Characterized by genuine multi-path interference and severe hardware noise under low signal-to-background ratios (SBR), this dataset serves as a strict zero-shot evaluation benchmark to verify the model's effectiveness in resolving monocular scale ambiguity under real-world scenarios.

\subsection{Evaluation Metrics}
\label{subsec:metrics}
To quantitatively assess the performance of depth estimation models, we adopt six standard and widely accepted metrics following the evaluation protocol established in \cite{eigen2014depth}. Let $M$ denote the total number of valid pixels evaluated across the test dataset. For a specific pixel $i$, let $d_i$ and $\hat{d}_i$ represent the ground-truth depth value and the corresponding predicted depth value, respectively. The evaluation criteria are formally defined as follows:

\begin{itemize}
	\item \textbf{Threshold Accuracy ($\delta_k$):} This metric computes the percentage of pixels whose maximum ratio between the predicted depth and the ground truth falls below a predefined threshold base $\tau = 1.25$:
	\begin{equation}
		\delta_k = \frac{1}{M} \sum_{i=1}^{M} \mathbb{I} \left( \max \left( \frac{d_i}{\hat{d}_i}, \frac{\hat{d}_i}{d_i} \right) < \tau^k \right)
		\label{eq:delta_k}
	\end{equation}
	where $\mathbb{I}(\cdot)$ denotes the indicator function. Following standard practice, we report the threshold accuracy at three progressive levels: $\delta_1$ ($\tau < 1.25$), $\delta_2$ ($\tau < 1.25^2$), and $\delta_3$ ($\tau < 1.25^3$). Higher values ($\uparrow$) indicate a more precise local alignment.
	
	\item \textbf{Absolute Relative Error ($\text{Abs Rel}$):} This error index measures the average relative deviation by normalizing the absolute pixel-wise discrepancy with the ground-truth depth value:
	\begin{equation}
		\text{Abs Rel} = \frac{1}{M} \sum_{i=1}^{M} \frac{\left| d_i - \hat{d}_i \right|}{d_i}
		\label{eq:abs_rel}
	\end{equation}
	
	\item \textbf{Root Mean Squared Error ($\text{RMSE}$):} This metric evaluates the global structural alignment by squaring the residual errors, making it highly sensitive to large outliers:
	\begin{equation}
		\text{RMSE} = \sqrt{\frac{1}{M} \sum_{i=1}^{M} \left( d_i - \hat{d}_i \right)^2}
		\label{eq:rmse}
	\end{equation}
	
	\item \textbf{Mean $\log_{10}$ Error ($\log_{10}$):} To mitigate the scale dominance of distant pixels and focus on the logarithmic error scale, the average absolute difference in the $\log_{10}$ space is formulated as:
	\begin{equation}
		\log_{10} = \frac{1}{M} \sum_{i=1}^{M} \left| \log_{10} d_i - \log_{10} \hat{d}_i \right|
		\label{eq:log10}
	\end{equation}
\end{itemize}

For all error-based metrics including $\text{Abs Rel}$ \eqref{eq:abs_rel}, $\text{RMSE}$ \eqref{eq:rmse}, and $\log_{10}$ \eqref{eq:log10}, lower values ($\downarrow$) denote superior depth estimation accuracy and higher structural fidelity.

\subsection{Implementation Details}
We set the number of optimization steps to $T=2$ with a learning rate of $\eta = 0.002$ for updating our HSN framework across all datasets. 
As detailed in Section~\ref{subsec:Statistical Distribution Alignment}, two strategies are employed to handle in-domain and unseen data, respectively, each featuring its own distinct distribution alignment and adaptive constraint modulation suite $\mathcal{P} = \{\lambda_f, \gamma, \epsilon\}$, unless otherwise mentioned in the model configuration.
For in-domain experiments, the models are pre-trained on the training split of the target domain prior to evaluation. 
Conversely, in zero-shot experiments, we evaluate the models directly on entirely unseen target domains without any domain-specific fine-tuning. All comparative depth networks follow their original practices and settings. HSN is applicable to a wide range of MDE networks with encoder–decoder architectures. To validate the generality of HSN, we integrate it with depth estimation networks of various architectures and scales, including the relatively lightweight AdaBins~\cite{bhat2021adabins}, the RGB+SPAD multimodal PhoMoNet~\cite{lee2023monocular}, and large-scale pretrained models such as ZoeD-M12~\cite{bhat2023zoedepth}, DAV1~\cite{yang2024depth}, DAV2~\cite{yang2024depthv2}, DepthPro~\cite{bochkovskiydepth}, and UniDepthV2~\cite{piccinelli2025unidepthv2}. "N" and "H" indicate models fine-tuned on NYU Depth V2~\cite{eigen2014depth} and Hypersim~\cite{roberts2021hypersim}, while "S", "B", and "L" represent the Small, Base, and Large backbone scales, respectively. We conduct the experiments using PyTorch on a single NVIDIA RTX 3090 GPU, whereas baseline results are either locally reproduced using their official code or directly adopted from the original literature.

\subsection{Comparison with State-of-the-Art}

\begin{table}[t]
	\centering
	\caption{Quantitative in-domain comparison on NYU Depth V2. \textsuperscript{\textit{re}} indicates our re-evaluation using the official codes and same environments as our method for a fair TTA comparison. "N" indicates that the model is fine-tuned on NYU Depth V2, and "L" denotes the Large backbone version. \textbf{Bold} and \underline{underline} indicate the best and second best results respectively.}
	\label{tab:indomain_nyu_results}
	\resizebox{\columnwidth}{!}{
		\begin{tabular}{l ccc ccc}
			\toprule
			\textbf{Method} & {\boldmath$\delta_1$} $\uparrow$ & {\boldmath$\delta_2$} $\uparrow$ & {\boldmath$\delta_3$} $\uparrow$ & \textbf{Abs Rel} $\downarrow$ & \textbf{RMSE} $\downarrow$ & {\boldmath$\log_{10}$} $\downarrow$ \\ 
			\midrule
			\multicolumn{4}{l}{\textbf{Baselines}} \\
			Eigen \etal~\cite{eigen2014depth}     & 0.769 & 0.950 & 0.988 & 0.158 & 0.641 & – \\
			DORN~\cite{fu2018deep}            & 0.828 & 0.965 & 0.992 & 0.115 & 0.509 & 0.051 \\
			Hu \etal~\cite{hu2019revisiting}        & 0.866 & 0.975 & 0.993 & 0.115 & 0.530 & 0.050 \\
			BTS~\cite{lee2019big}              & 0.885 & 0.978 & 0.994 & 0.110 & 0.392 & 0.047 \\
			LocalBins~\cite{bhat2022localbins}         & 0.907 & 0.987 & 0.998 & 0.099 & 0.357 & 0.042 \\
			NeWCRFs~\cite{yuan2022neural}          & 0.922 & 0.992 & 0.998 & 0.095 & 0.334 & 0.041 \\
			DepthFormer(LR)~\cite{li2023depthformer}          & 0.923 & 0.989 & 0.997 & 0.094 & 0.329 & 0.040 \\
			DST(w/DenseDepth)~\cite{nishimura2020disambiguating}           & 0.926 & 0.983 & 0.995 & 0.081 & 0.346 & 0.035 \\
			ScaleDepth-N~\cite{zhu2024scaledepth}           & 0.957 & 0.994 & \underline{0.999} & 0.074 & 0.267 & 0.032 \\
			DAV2-N-L~\cite{yang2024depthv2}           & \underline{0.984} & \underline{0.998} &  \textbf{1.000} & 0.056 & 0.206 & \underline{0.024} \\
			SPADnet(Linear)~\cite{sun2020spadnet}           & 0.970 & 0.993 & 0.998 & \underline{0.031} &  0.347 & – \\
			SPADnet(Log)~\cite{sun2020spadnet}           & \textbf{0.996} &  \textbf{0.999} & \underline{0.999} & \textbf{0.010} & \textbf{0.144} & – \\
			\midrule
			\multicolumn{4}{l}{\textbf{Ours}} \\
			AdaBins\textsuperscript{\textit{re}}~\cite{bhat2021adabins}          & 0.901 & 0.983 & 0.997 & 0.104 & 0.366 & 0.044 \\
			AdaBins-HSN          & 0.903 & 0.983 & 0.997 & 0.102 & 0.364 & 0.044 \\
			PhoMoNet\textsuperscript{\textit{re}}~\cite{lee2023monocular}           & 0.952 & 0.991 & 0.997 & 0.073 & 0.287 & 0.031 \\
			PhoMoNet-HSN      & 0.952 & 0.991 & 0.997 & 0.073 & 0.287 & 0.031 \\
			ZoeD-M12-N\textsuperscript{\textit{re}}~\cite{bhat2023zoedepth}           & 0.954 & 0.995 & \underline{0.999} & 0.075 & 0.270 & 0.032 \\
			ZoeD-M12-N-HSN           & 0.957 & 0.995 & \underline{0.999} & 0.071 & 0.264 & 0.031 \\
			DAV1-N-L\textsuperscript{\textit{re}}~\cite{yang2024depth}           & \underline{0.984} & \underline{0.998} & \textbf{1.000} & 0.056 & 0.206 & \underline{0.024} \\
			DAV1-N-L-HSN           & \underline{0.984} & \underline{0.998} & \textbf{1.000} & 0.052 & \underline{0.200} & \textbf{0.023} \\
			\bottomrule
		\end{tabular}
	}
\end{table}

\begin{table*}[t]
	\centering
	\setlength{\tabcolsep}{3pt}
	\scriptsize
	
	\begin{minipage}{0.48\linewidth}
		\centering
		\caption{Zero-shot performance comparison on NYU Depth V2 dataset. \textbf{Bold} and \underline{underline} indicate the best and second best results respectively.}
		\label{tab:zeroshot_nyu_results}
		\begin{tabular}{lccc}
			\toprule
			\textbf{Method} & $\boldsymbol{\delta_1} \uparrow$ & \textbf{Abs Rel $\downarrow$} & \textbf{RMSE $\downarrow$} \\
			\midrule
			
			\multicolumn{4}{l}{\textbf{Baseline Zero-shot}} \\
			HG-Depth~\cite{zhang2020unsupervised} & 0.350 & 0.513 & 1.457 \\
			DepthCLIP~\cite{zhang2022can} & 0.394 & 0.388 & 1.167 \\
			DAV2-H-S~\cite{yang2024depthv2} & 0.700 & 0.208 & 0.509 \\
			DAV2-H-B~\cite{yang2024depthv2} & 0.704 & 0.205 & 0.510 \\
			DAV2-H-L~\cite{yang2024depthv2} & 0.714 & 0.201 & 0.505 \\
			ZeroDepth~\cite{guizilini2023towards} & 0.926 &  0.081 & 0.338 \\
			
			Metric3D~\cite{yin2023metric3d} & 0.944 & 0.083 &  0.310 \\
			Metric3D V2~\cite{hu2024metric3d} &  0.975 & 0.063 &  0.251 \\
			DepthPro~\cite{bochkovskiydepth} & 0.942  & 0.090 & 0.265  \\
			UniDepthV2-S~\cite{piccinelli2025unidepthv2} & 0.944  & 0.085 & 0.240  \\
			UniDepthV2-B~\cite{piccinelli2025unidepthv2} & 0.956 & 0.077 & 0.219  \\
			UniDepthV2-L~\cite{piccinelli2025unidepthv2} & 0.971 & 0.063 & 0.198  \\
			UniDepthV2-N-L~\cite{piccinelli2025unidepthv2} &  \textbf{0.988} & 0.047 &  \textbf{0.180} \\
			
			\midrule
			
			\multicolumn{4}{l}{\textbf{TTA Zero-shot}} \\
			MfH~\cite{zhao2024metric} &  0.832 & 0.137 & 0.487 \\
			DR-LiDAR1~\cite{marsal2025simple} & 0.939 & 0.063 &  0.652 \\
			DR-LiDAR16~\cite{marsal2025simple} & 0.976 &  \textbf{0.039} & 0.454 \\
			DR-LiDAR32~\cite{marsal2025simple} &  0.974 & \underline{0.040} & 0.461 \\
			
			\midrule
			
			\multicolumn{4}{l}{\textbf{Ours}} \\
			
			DAV2-H-S-HSN & 0.796 & 0.149 & 0.391 \\
			DAV2-H-B-HSN & 0.827 & 0.139 & 0.368 \\
			DAV2-H-L-HSN & 0.811 & 0.147 & 0.370 \\
			DepthPro-HSN & 0.948 & 0.084 & 0.255 \\
			UniDepthV2-S-HSN & 0.961 & 0.075 & 0.230 \\
			UniDepthV2-B-HSN & 0.973 & 0.064 & 0.200 \\
			UniDepthV2-L-HSN & \underline{0.980} & 0.056 & \underline{0.186} \\

			\bottomrule
		\end{tabular}
	\end{minipage}
	\hfill
	\begin{minipage}{0.48\linewidth}
		\centering
		\caption{Zero-shot evaluation on real-world RGB-SPAD dataset. \textbf{Bold} and \underline{underline} indicate the best and second best results respectively.}
		\label{tab:zeroshot_rgb_spad_results}
		\begin{tabular}{lccc}
			\toprule
			\textbf{Method} & $\boldsymbol{\delta_1} \uparrow$ & \textbf{Abs Rel $\downarrow$} & \textbf{RMSE $\downarrow$} \\
			\midrule
			\multicolumn{4}{l}{\textbf{Baselines}} \\
			DepthFormer(MS)~\cite{agarwal2022depthformer} & 0.242 & 0.596 & 1.450 \\
			AdaBins~\cite{bhat2021adabins} & 0.424 & 0.347 & 1.103 \\
			PhoMoNet~\cite{lee2023monocular} & 0.840 & \underline{0.154} & \underline{0.508} \\
			ZoeD-M12-N~\cite{bhat2023zoedepth} & 0.842 & 0.216 & 0.518 \\
			DAV1-N-L~\cite{yang2024depth} & \textbf{0.879} & 0.232 & 0.529 \\
			DAV2-H-L~\cite{yang2024depthv2} & 0.757 & 0.770 & 1.195 \\
			DAV2-H-B~\cite{yang2024depthv2} & 0.676 & 0.726 & 1.186 \\
			DAV2-H-S~\cite{yang2024depthv2} & 0.751 & 0.286 & 0.621 \\
			
			\midrule
			\multicolumn{4}{l}{\textbf{Ours}} \\
			AdaBins-HSN & 0.689 & 0.282 & 0.842 \\
			PhoMoNet-HSN & \underline{0.860} & \textbf{0.141} & \textbf{0.474} \\
			ZoeD-M12-N-HSN & 0.842 & 0.213 & 0.513 \\
			DAV1-N-L-HSN & \textbf{0.879} & 0.231 & 0.527 \\
			DAV2-H-L-HSN & 0.756 & 0.571 & 1.049 \\
			DAV2-H-B-HSN & 0.691 & 0.592 & 1.064 \\
			DAV2-H-S-HSN & 0.773 & 0.214 & 0.543 \\
			
			\bottomrule
		\end{tabular}
	\end{minipage}
	
\end{table*}

{\bf In-domain comparison on NYU Depth V2:} We compare our four depth model variants equipped with HSN to 16 benchmark models, which involves three mainstream categories in monocular depth estimation. For lightweight networks with specific structural designs, we evaluate against Eigen \etal~\cite{eigen2014depth}, DORN~\cite{fu2018deep}, Hu \etal~\cite{hu2019revisiting}, BTS~\cite{lee2019big}, NeWCRFs~\cite{yuan2022neural}, AdaBins~\cite{bhat2021adabins}, and LocalBins~\cite{bhat2022localbins} that focus on computational efficiency and discrete bin optimization. For Transformer-based and large vision foundation models, we compare with DepthFormer(LR)~\cite{li2023depthformer}, ScaleDepth-N~\cite{zhu2024scaledepth}, ZoeD-M12-N~\cite{bhat2023zoedepth}, and DAV1~\cite{yang2024depth}/DAV2~\cite{yang2024depthv2}-N-L that leverage large backbones for open-world generalization. For sensor-fused networks, we include SPADnet (Linear/Log)~\cite{sun2020spadnet} and PhoMoNet~\cite{lee2023monocular} which explicitly resolve scale ambiguity by coupling RGB imagery respectively with 3D dense or 1D sparse physical SPAD data. This extensive evaluation across diverse architectures validates the efficacy and compatibility of our proposed HSN framework as a plug-and-play solution for TTA.

As summarized in Table~\ref{tab:indomain_nyu_results}, several vital observations can be derived from the quantitative results. First, our HSN framework demonstrates robust architectural adaptability, consistently yielding stable performance gains across three out of the four baseline networks (\ie, AdaBins, ZoeD-M12-N, and DAV1-N-L). Note that for PhoMoNet, since it already explicitly feeds 1D active SPAD measurements for deep feature fusion with 2D images during the full end-to-end training phase, its baseline weights heavily encode the 1D physical information, thereby saturating the spatial layout and leaving negligible margin for subsequent HSN-driven geometric optimization during inference. Second, our framework achieves top-tier SOTA dominance in terms of overall metrics. Specifically, our strongest variant (DAV1-N-L-HSN) achieves a superior RMSE of 0.200, surpassing 15 out of the 16 competitive SOTA networks. Meanwhile, our second-strongest variant (ZoeD-M12-N-HSN) records an RMSE of 0.264, outperforming 13 out of 16 SOTA baselines and notably beating advanced architectures such as ScaleDepth-N. Finally, we notice a crucial performance advantage over the 3D dense network with RGB-SPAD sensor fusion. Compared against the 3D dense SPADnet (Linear), our TTA design of 1D SPAD guidance effectively suppresses fatal large distance estimation errors, slashing the RMSE metric by a massive 42.4\% (0.200 \vs 0.347). This remarkable gap highlights that our framework can secure much higher absolute metric precision while utilizing active SPAD sensing signals of significantly lower dimensions.

{\bf Zero-shot generalization on NYU Depth V2:} Our seven HSN variants are evaluated against 17 zero-shot depth benchmarks, consisting of one CNN model (HG-Depth~\cite{zhang2020unsupervised}), 12 depth foundation models, and four recent TTA counterparts. Additionally, UniDepthV2-N-L~\cite{piccinelli2025unidepthv2} (fine-tuned on the target domain) serves as a performance upper bound (Oracle). As reported in Table~\ref{tab:zeroshot_nyu_results}, HSN yields universal, synchronous improvements across all metrics and architectures. Specifically, for depth foundation models, HSN shatters their zero-shot performance ceilings; e.g., it pushes DepthPro~\cite{bochkovskiydepth} to $\delta_1 = 0.948$ and UniDepthV2-L~\cite{piccinelli2025unidepthv2} to a remarkable $\delta_1 = 0.980$, achieving an 11.1\% relative error reduction in Abs Rel. On lighter baselines like DAV2-H-B~\cite{yang2024depthv2}, HSN delivers a massive 17.5\% absolute gain in $\delta_1$ and cuts the Abs Rel by 32.2\%. 

Crucially, as a TTA framework, HSN offers superior efficiency over existing counterparts. Unlike MfH~\cite{zhao2024metric}'s heavy diffusion computation or DR-LiDAR~\cite{marsal2025simple}'s reliance on dense 3D point clouds, HSN maintains a lightweight design and operates under extremely sparse regimes by utilizing only 1D SPAD signals. Despite HSN's compact architecture and data sparsity, our advanced variants (e.g., UniDepthV2-B/L-HSN) competitively match or outperform these heavy TTA methods, particularly across $\delta_1$ and RMSE indicators. Strikingly, without any target-domain training, our UniDepthV2-L-HSN ($\delta_1 = 0.980$, $\text{RMSE} = 0.186$) closely approaches the fully-supervised Oracle ($\delta_1 = 0.988$, $\text{RMSE} = 0.180$), demonstrating its immense practical benefit for zero-shot metric depth estimation. Notably, this evaluation reveals that HSN brings significantly more pronounced improvements to zero-shot scenarios than to in-domain settings, firmly validating that HSN primarily excels at bridging cross-domain generalization gaps rather than merely providing marginal refinements to in-domain setups.

\textbf{Real-world evaluation on RGB-SPAD Dataset:} To assess the zero-shot performance of our HSN towards real-world indoor scenes, we conduct extensive TTA experiments across our seven variants compared to eight MDE baselines. 
Table~\ref{tab:zeroshot_rgb_spad_results} shows that without any training on the target domain, our HSN framework consistently yields notable performance gains across almost all architectures. 
Specifically, PhoMoNet-HSN achieves the state-of-the-art (SOTA) overall performance, significantly reducing the Abs Rel from $0.154$ to $0.141$ and the RMSE from $0.508$ to $0.474$. 
Meanwhile, DAV1-N-L-HSN maintains the highest $\delta_1$ accuracy of $0.879$ while suppressing the Abs Rel and RMSE. 
Notably, HSN effectively boosts PhoMoNet~\cite{lee2023monocular} in unseen realistic scenarios relative to its in-domain setting, allowing it to even outperform the acknowledged strong baseline DAV1-N-L~\cite{yang2024depth} in terms of overall error metrics. 
Crucially, our HSN framework exhibits substantial improvements on weaker baselines (e.g., AdaBins~\cite{bhat2021adabins} and the DAV2-H family~\cite{yang2024depthv2}), demonstrating its remarkable robustness. 
These results strongly prove that HSN serves as a generic, plug-and-play TTA framework that effectively mitigates domain shifts on the fly and unlocks the zero-shot potential of diverse architectures in real-world applications.

\begin{figure*}[htbp]
	\centering
	\includegraphics[width=0.48\textwidth]{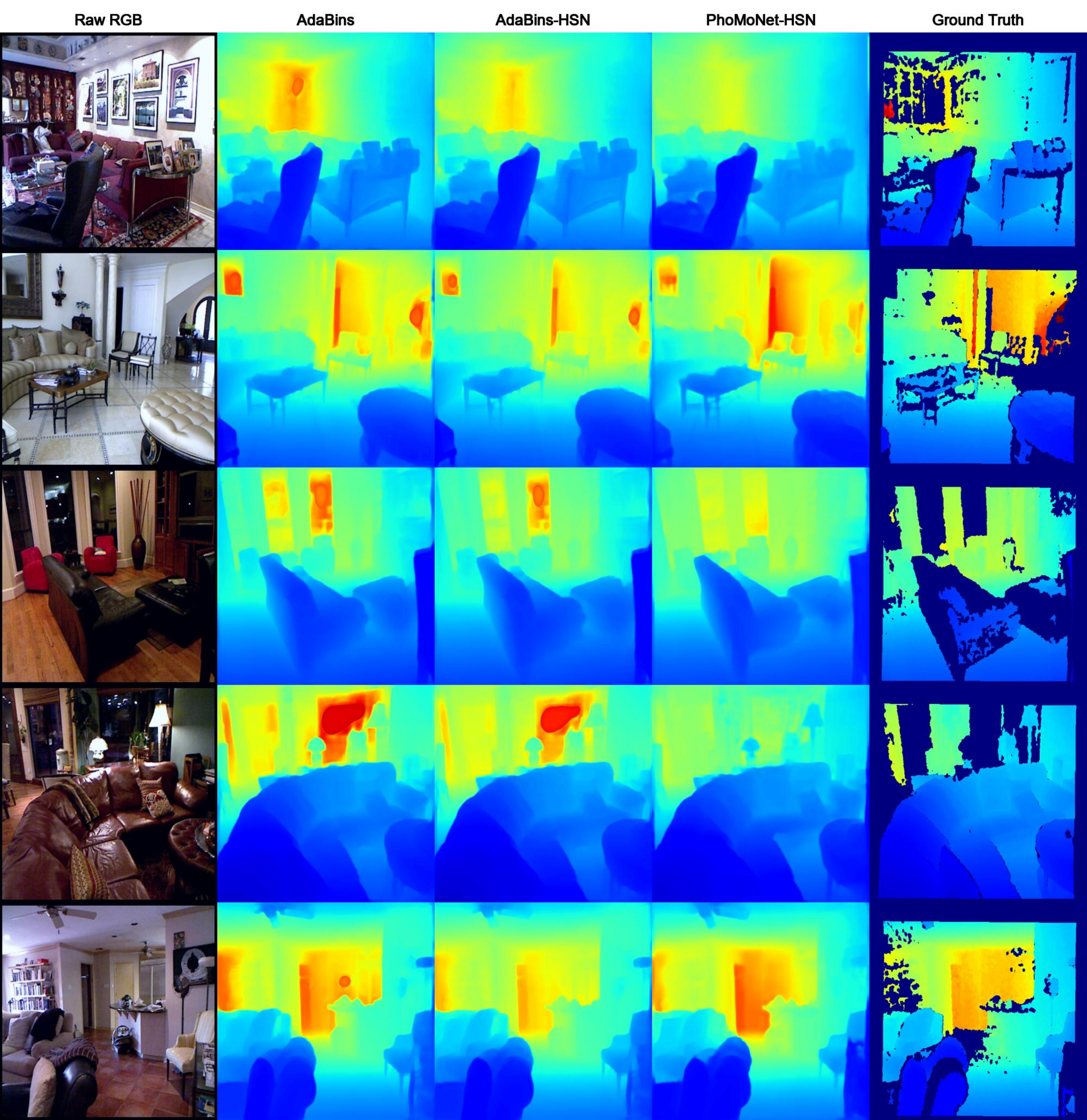}
	\hfill
	\includegraphics[width=0.48\textwidth]{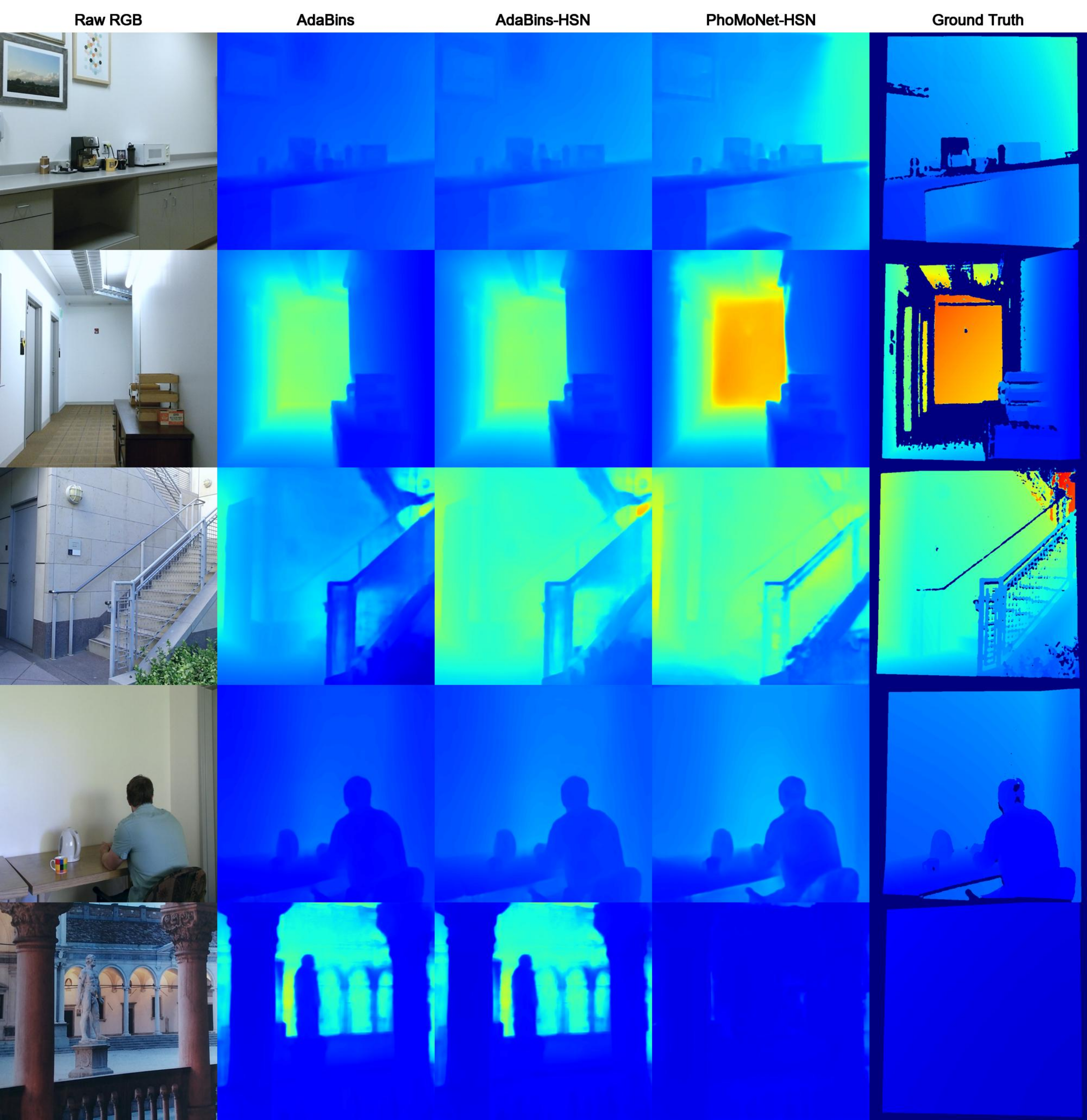}
	\caption{Comparison on the NYU Depth V2 (left) and RGB-SPAD (right) datasets. Only the last sample on the right is a poster; all other samples are captured scenes.}
	\label{fig:visual_comparison}
\end{figure*}

\textbf{Visual analysis:} As illustrated in Fig.~\ref{fig:visual_comparison} (left), AdaBins-HSN effectively suppresses all the erroneous depth artifacts (the red regions) that trouble its baseline counterpart, and even surpasses PhoMoNet-HSN in challenging scenarios, such as those in Row 2 and Row 5. Nevertheless, from an overall perspective, PhoMoNet-HSN exhibits a closer alignment with the ground truth than the other two models due to its inherent cross-modal feature fusion.

Regarding Fig.~\ref{fig:visual_comparison} (right), AdaBins-HSN addresses most of the ambiguous boundary transitions and scale issues that its baseline suffers from. Furthermore, PhoMoNet-HSN outperforms them by showing sharper structural edges and more consistent global depth layouts. It is worth noting that PhoMoNet-HSN is the only model that predicts a depth substantially close to the planar ground truth for the most challenging poster in Row 5, whereas other methods are misled by its rich objects and textures.

\subsection{Ablation Study}
\label{subsec:Ablation Studies}

\begin{table}[htbp]
	\centering
	\caption{Effect of HSN components on AdaBins inference. Adabins is trained on NYU-Depth-v2 (Train). In-Domain performance is evaluated on NYU-Depth-v2 (Test), while zero-shot generalization (Unseen) is evaluated on the RGB-SPAD dataset. \textbf{Bold} and \underline{underline} indicate the best and second best results respectively.}
	\label{tab:HSN_components}
	\setlength{\tabcolsep}{3.5pt} 
	\resizebox{\columnwidth}{!}{%
		\begin{tabular}{lcccccc}
			\toprule
			\multirow{2}{*}{\textbf{Modification}} & \multicolumn{3}{c}{\textbf{In-Domain}} & \multicolumn{3}{c}{\textbf{Unseen}} \\ \cmidrule(r){2-4} \cmidrule(l){5-7}
			& {\boldmath $\delta_1$} $\uparrow$ & \textbf{Abs Rel $\downarrow$} & $\textbf{RMSE} \downarrow$ & {\boldmath $\delta_1$} $\uparrow$ & $\textbf{Abs Rel} \downarrow$ & $\textbf{RMSE} \downarrow$ \\ \midrule
			Baseline (Original)             & 0.901 & 0.104 & 0.366          &  0.424 & 0.347 & 1.103          \\
			w/o $\omega$            & \underline{0.903} & \underline{0.102} & \underline{0.364}          & \underline{0.688} & \textbf{0.282} & \underline{0.842}    \\
			only $\mathcal{L}_{phy}$            & \textbf{0.906} & \textbf{0.101} & \textbf{0.361}          & 0.687 & 0.284 & \textbf{0.841}          \\
			only $\mathcal{L}_{fid}$      & \underline{0.903} & 0.103 & 0.366   & 0.687 & \underline{0.283} & \textbf{0.841}          \\
			w/o ZoDA             & - & - & -          & 0.427 & 0.332 & 1.080          \\
			w/o SA              & 0.879 & 0.119 & 0.413          & 0.458 & 0.339 & 1.075          \\ \midrule
			\textbf{Full Proposed} & \underline{0.903} & \underline{0.102} & \underline{0.364} & \textbf{0.689} & \textbf{0.282} & \underline{0.842} \\ \bottomrule
		\end{tabular}%
	}
\end{table}

{\bf Effect of HSN components:} To investigate the effect of different components, we conduct an extensive ablation study categorized into loss formulations and structural modules, as shown in Table~\ref{tab:HSN_components}. Regarding the loss variants (\text{w/o } $\omega$ to only $\mathcal{L}_{fid}$), removing the noise-aware control (\text{w/o } $\omega$) impacts cross-domain robustness, while evaluating the individual sub-losses reveals their distinct characteristics. Notably, while only $\mathcal{L}_{phy}$ yields the optimal in-domain precision ($\delta_1 = 0.906$), both sub-losses independently boost zero-shot generalization (driving the unseen $\delta_1$ from 0.424 to 0.687) and exhibit a complementary synergy that peaks at 0.689 in our full method. Regarding structural components, the elimination of either the Zero-order Distribution Alignment (ZoDA) or the Stability Audit Mechanism (SA) incurs severe performance degradation, particularly in zero-shot generalization. For instance, removing ZoDA causes a substantial drop in the unseen $\delta_1$ (down to 0.427), demonstrating its critical role in cross-domain transfer, while stripping SA leads to noticeable accuracy drops across both testing scenarios. Ultimately, our full proposed method strikes a better balance, unlocking superior cross-dataset generalization ($\delta_1 = 0.689$ on RGB-SPAD) while preserving highly comparable in-domain performance.

\begin{table}[t]
	\centering
	\caption{Cross-strategy evaluation of data domains using AdaBins. We evaluate the compatibility between domain-specific strategies and data distributions across multiple depth metrics. Bold values indicate optimal matching, and arrows ($\uparrow$ for $\delta_1$, $\downarrow$ for others) denote the preferred metric directions, where performance degradation occurs under strategy-data mismatch.}
	\label{tab:cross_strategy}
	\resizebox{\linewidth}{!}{%
		\begin{tabular}{@{}lcccccc@{}}
			\toprule
			\textbf{Strategy} & \multicolumn{3}{c}{\textbf{In-Domain data}} & \multicolumn{3}{c}{\textbf{Unseen data}} \\ \cmidrule(lr){2-4} \cmidrule(lr){5-7}
			& {\boldmath $\delta_1$} $\uparrow$ & \textbf{Abs Rel $\downarrow$} & \textbf{RMSE $\downarrow$} & {\boldmath $\delta_1$} $\uparrow$ & \textbf{Abs Rel $\downarrow$} & \textbf{RMSE $\downarrow$} \\ \midrule
			Baseline (Original)             & 0.901 & 0.104 & 0.366          &  0.424 & 0.347 & 1.103          \\ \midrule
			In-Domain Strategy & \textbf{0.903} & \textbf{0.102} & \textbf{0.364} & 0.427  & 0.333  & 1.081  \\
			Unseen Strategy  & 0.811  & 0.129  & 0.436  & \textbf{0.689} & \textbf{0.282} & \textbf{0.842} \\ \bottomrule
	\end{tabular}}
\end{table}

{\bf Necessity of Domain-Specific Data Strategies:} To validate the necessity of our domain-specific configurations, we conduct a cross-strategy evaluation by interchanging two schemes of distribution alignment and adaptive constraint modulation $\mathcal{P}$ between the in-domain and unseen settings. Crucially, these two strategies differ in their design: the in-domain strategy disables ZoDA while employing its refinement set $\mathcal{P}_{\text{in}}=\{\lambda_f, \gamma, \epsilon\}$, whereas the unseen domain strategy activates ZoDA combined with a conservative set $\mathcal{P}_{\text{un}}=\{\lambda_f, \gamma, \epsilon\}$. As observed in Table~\ref{tab:cross_strategy}, enforcing these mismatched strategy configurations onto opposite domains leads to severe performance degradation. The results demonstrate that while the unseen strategy significantly boosts performance in unseen scenarios by bridging the global domain bias ($\delta_1 = 0.689$), applying it to in-domain data leads to over-correction, crashing the in-domain $\delta_1$ to 0.811. Conversely, forcing the in-domain strategy on unseen data fails to generalize ($\delta_1 = 0.427$). This distinct performance asymmetry firmly justifies that both the conditional utilization of ZoDA and their respective adaptive parameter sets are uniquely and optimally tailored to their target distributions.

\begin{figure}[t] 
	\centering
	\begin{subfigure}[b]{0.49\linewidth}
		\centering
		\includegraphics[width=\linewidth]{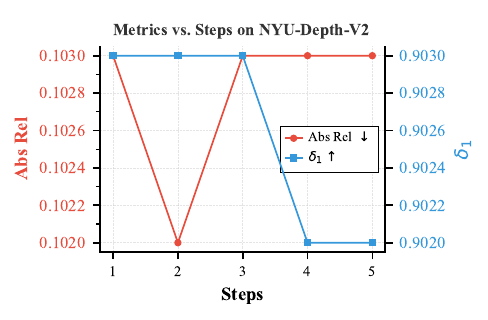}
		\label{fig:Metrics VS Steps on NYU-Depth-V2}
	\end{subfigure}
	\hfill 
	\begin{subfigure}[b]{0.49\linewidth}
		\centering
		\includegraphics[width=\linewidth]{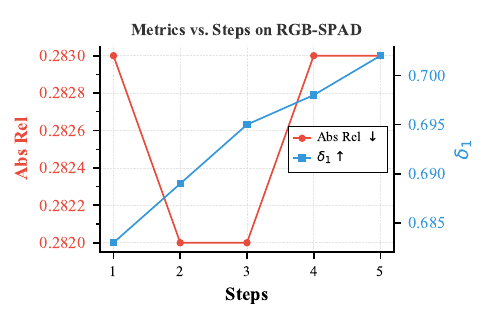}
		\label{fig:Metrics VS Steps on RGB-SPAD}
	\end{subfigure}
	\caption{Metric curves vs. optimization steps for NYU-Depth-V2 (left) and RGB-SPAD (right). We apply our method on AdaBins to observe the performance across 1 to 5 optimization steps.}
	\label{fig:steps_metrics}
\end{figure}

{\bf Effect of optimization steps:} We apply HSN to AdaBins to investigate its in-domain and zero-shot performance across 1 to 5 optimization steps. Fig.~\ref{fig:steps_metrics} (left) demonstrates that optimal performance for in-domain inference is achieved at 2 steps, while Fig.~\ref{fig:steps_metrics} (right) indicates that 3 steps provide the best trade-off for unseen domains. This discrepancy suggests that zero-shot adaptation inherently requires more optimization to align unfamiliar feature distributions compared to in-domain inference. Considering both accuracy and computational efficiency, we choose 2 optimization steps for our default setting, unless otherwise specified.

\section{Conclusion}
\label{sec:conclusion}
In this work, we have established a novel noise-driven paradigm for monocular metric depth estimation by shifting the perception of SPAD noise from a pure corruption to a valuable environmental prior. Through the development of our HSN framework, we demonstrated that the native noise signal can be naturally exploited via GA modules to enforce accurate geometric and scale calibration during inference, thereby bypassing the severe latency overhead or dense 3D data dependencies that typically bottleneck competitive TTA counterparts in resource-constrained platforms. The proposed test-time gated attention mechanism performs dynamic attention allocation across hierarchical encoder feature representations in an interpretable manner that aligns with the theories of hierarchical representation robustness.

Our extensive experiments on real-world indoor benchmarks featuring synthetic and realistic SPAD histograms validate the efficacy of this framework, showcasing not only state-of-the-art in-domain precision but also remarkable adaptability under zero-shot generalization to unfamiliar environments. Moving forward, this work highlights the untapped potential of raw sensor noise in sparse data regimes, opening new possibilities for building highly efficient, multi-modal perception systems across broader applications such as robotics, AR, and autonomous navigation.
{
    \small
    \bibliographystyle{ieeenat_fullname}
    \bibliography{main}
}

\end{document}